%% file: cas-refs.tex
\documentclass[preprint,12pt]{elsarticle}
\usepackage{graphicx}

\input{math_commands.tex}

\usepackage{hyperref}
\usepackage{url}
\usepackage{booktabs} 
\usepackage{color}

\usepackage{mathtools}
\usepackage{amsmath}
\usepackage{multirow}
\usepackage[ruled,vlined]{algorithm2e}
\usepackage{subfig}
\usepackage{wrapfig}
\usepackage{textcomp}

\usepackage{amssymb}

\journal{Artificial Intelligence}

\begin{document}

\begin{frontmatter}



\title{Multimodal Variational Autoencoders for Semi-Supervised Learning: In Defense of Product-of-Experts
}


\affiliation[inst1]{organization={Novo Nordisk Foundation Center for Biosustainability, Technical University of Denmark},
            addressline={Kemitorvet 220}, 
            city={Copenhagen},
            postcode={2800}, 
            country={Denmark}}
            
\affiliation[inst2]{organization={Department of Computer Science, University of Copenhagen},
            addressline={Universitetsparken 1}, 
            city={Copenhagen},
            postcode={2100}, 
            country={Denmark}}
            
\author[inst1]{Svetlana Kutuzova}
\author[inst2]{Oswin Krause}
\author[inst1]{Douglas McCloskey}
\author[inst2]{Mads Nielsen}
\author[inst2]{Christian Igel}

\begin{abstract}
Multimodal generative models should be able to learn a meaningful latent representation that enables a coherent joint generation of all modalities (e.g., images and text). Many applications also require the ability to accurately sample modalities conditioned on observations of a subset of the modalities. Often not all modalities may be observed for all training data points, so semi-supervised learning should be possible.
In this study, we propose a novel product-of-experts (PoE) based variational autoencoder that have these desired properties. We benchmark it against a mixture-of-experts (MoE) approach and a PoE approach of combining the modalities with an additional encoder network. An empirical evaluation shows that the PoE based models can outperform the contrasted models. 
Our experiments support the intuition that PoE models are more suited for a conjunctive combination of modalities. 
\end{abstract}




\end{frontmatter}

\DeclarePairedDelimiterX{\infdivx}[2]{(}{)}{%
  #1\;\delimsize\|\;#2%
}
\newcommand{\kld}[2]{\ensuremath{D_{\text{KL}}\infdivx{#1}{#2}}\xspace}
\newcommand{\Lagr}{\mathcal{L}}
\newcommand{\norm}[1]{\left\lVert #1 \right\rVert}
\renewcommand{\d}{\text{d}}

\section{Introduction}
Multimodal generative modelling is important because information about real-world objects typically comes in different representations, or modalities. The information provided by each modality  may be erroneous and/or  incomplete, and  a complete reconstruction of the full information can  often only be achieved  by combining several modalities. For example, in image- and video-guided translation \citep{Caglayan20194159},  additional visual context can potentially resolve ambiguities (e.g., noun genders) when translating  written text. 

In many applications,  modalities may be missing for a subset of the observed samples during training and deployment. Often the description of an object in one modality is easy to obtain, while annotating it with another modality is slow and expensive. Given two modalities, we call samples \emph{paired}  when both modalities are present, and \emph{unpaired} if one is missing. 
The simplest way to deal with paired and unpaired training examples is to discard the unpaired observations for learning. The smaller the share of paired samples, the more important becomes the ability to additionally learn from the unpaired data, referred to as \emph{semi-supervised learning} in this context (following the terminology from \citealp{VAEVAE}. Typically one would associate semi-supervised learning with learning form labelled and unlabelled data to solve a classification or regression tasks). Our goal is to provide a model that can leverage the information contained in unpaired samples and to investigate the capabilities of the model in  situations of low levels of supervision, that is, when only a few paired samples are available.
While a modality can be as low dimensional as a label, which can be handled by a variety of  discriminative models \citep{vanEngelen2020373}, we are interested in high dimensional modalities, for example  an image and a text caption. 

Learning a representation of multimodal data that allows to generate high-quality samples 
requires the following: 1) deriving a meaningful representation in a joint latent space for each high dimensional modality and 2) bridging the representations of different modalities in a way that the relations between them are preserved. The latter means that we do not want the modalities to be represented orthogonally in the latent space -- ideally the latent space should encode the object's properties independent of the input modality.
Variational autoencoders (\citealp{kingma2014autoencoding}) using a product-of-experts (PoE, \citealp{hinton2002training,Welling:2007}) approach for combining  input modalities are a promising approach for multimodal generative modelling having the desired properties, 
in particular the VAEVAE(a) model developed by  
\cite{NIPS2018_7801} and a novel model termed SVAE, which we present in this study. Both models can handle multiple high dimensional modalities, which  may not all be observed at  training time. 

It has been argued that a PoE approach is not well suited for multimodal generative modelling using variational autoencoders (VAEs) in comparison to  additive mixture-of-experts (MoE). 
It has empirically been shown that
the PoE-based MVAE   \citep{NIPS2018_7801} fails to properly model two high-dimensional modalities in  contrast to an (additive) MoE approach referred to as MMVAE, leading to the conclusion that ``PoE factorisation does not appear to be practically suited for multi-modal learning'' \citep{NIPS2019_9702}. 
This study sets out to test this conjecture for state-of-the-art multimodal VAEs.

The next section summarizes related work. Section~\ref{sec:svae} introduces SVAE as an alternative PoE based VAE approach derived from axiomatic principles. Then we present our experimental evaluation of multimodal VAEs before we conclude.

\section{Background and Related Work}\label{sec:background}
We consider multimodal generative modelling. 
We mainly restrict our considerations to two modalities $ x_1\in X_1, x_2 \in X_2$, where one modality may be missing at a time. Extensions to more modalities are discussed in Section~\ref{sec:svae-3}.
To address  the problem of generative cross-modal modeling, one modality $x_1$ can be generated from another modality $x_2$ by simply using independently trained generative models ($x_1 \rightarrow x_2$ and $x_2 \rightarrow x_1$) or a composed but non-interchangeable representation \citep{wang2016deep,NIPS2015_5775}. However, the ultimate goal of multimodal representation learning is to find a meaningful joint latent code distribution  bridging the two individual embeddings learned from $x_1$ and $x_2$ alone. This can be done by a two-step procedure that models the individual representations first and then applies an additional learning step to link them \citep{tian2019latent,silberer-lapata-2014-learning,Ngiam2011MultimodalDL}. In contrast, we focus on approaches that learn individual and joint representations simultaneously. 
Furthermore, our model should be able to learn in a semi-supervised setting. \cite{kingma2014semisupervised} introduced two models suitable for the case when one modality is high dimensional (e.g., an image) and another is low dimensional (e.g., a label) while our main interest are modalities of high complexity.

\begin{figure}
  \centering
  \includegraphics[width=.85\linewidth]{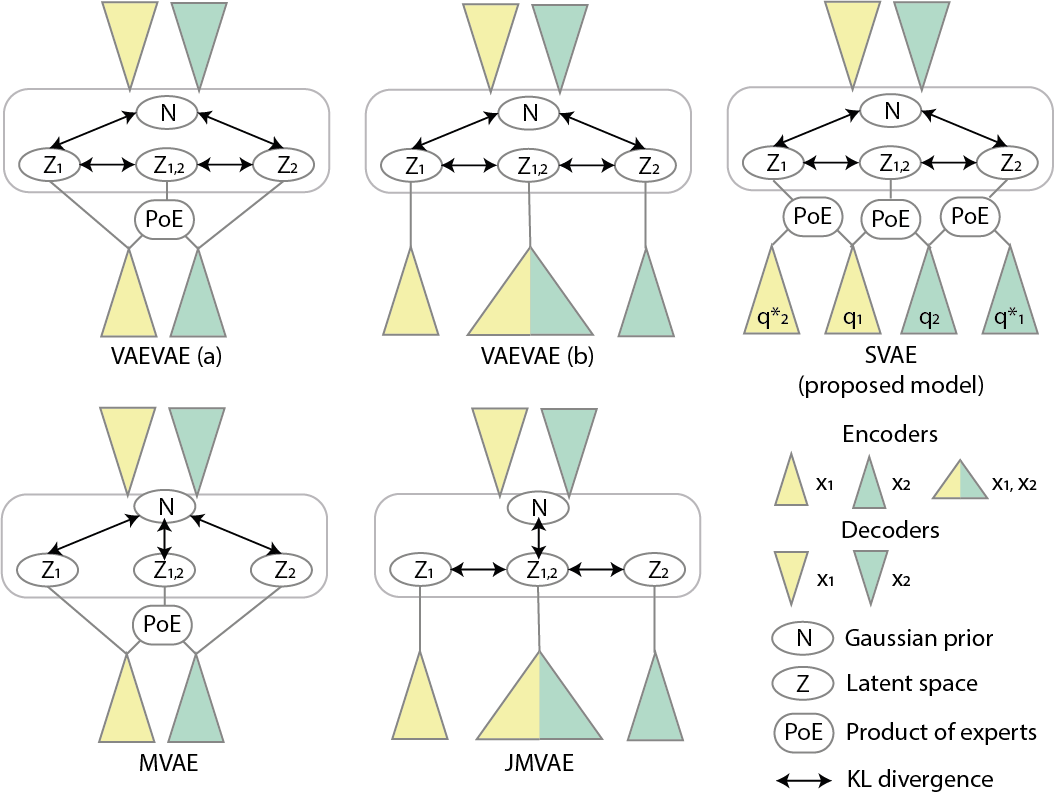}
  \caption{Schematic overview bi-modal VAEs using a  PoE and additional network structures that  are capable of  semi-supervised learning without requiring a two step learning procedure. VAEVAE (a) and (b) are by \cite{VAEVAE}, JMVAE is by \cite{2016arXiv161101891S}, MVAE is by \cite{NIPS2018_7801}, and SVAE is our newly proposed model. Each triangle stands for an individual neural network, the colors indicate the two different modalities.\label{fig:models}}
\end{figure}

We consider models based on variational autoencoders (VAEs, \citealp{kingma2014autoencoding,rezende2014stochastic}). Standard VAEs learn a latent representation $z \in Z$ for a set of observed variables $x \in X$ by modelling a joint distribution $p(x, z) = p(z)p(x | z)$. In the original VAE, the intractable posterior $q(z | x)$ and conditional distribution $p(x | z)$ are approximated by  neural networks trained by maximising the ELBO loss taking the form
\sloppy
\begin{equation}\label{eq_ELBO_single}
\Lagr = E_{q(z | x)}[\log{p(x | z)}] - \kld{q(z | x)}{\mathcal{N} (0, I)} 
\end{equation}
with respect to the parameters of the networks modelling $q(z | x)$ and  $p(x | z)$.
Here $\kld{\cdot}{\cdot}$ denotes the Kullback-Leibler divergence.
Bi-modal VAEs  that can handle a missing modality extend this approach by modelling $q(z|x_1, x_2)$ as well as $q_1(z|x_1)$ and $q_2(z|x_2)$, which replace the single  $q(z | x)$. Multimodal VAEs may 
differ  in 1) the way  they   approximate $q(z|x_1, x_2)$, $q_1(z|x_1)$ and $q_2(z|x_2)$ by neural networks and/or 2) the structure of the loss function,  see Figure~\ref{fig:models}.  Typically, there are no conceptual differences in the decoding, and we  model the decoding distributions in the same way for all methods considered in this study.

\cite{2016arXiv161101891S} introduced a model  termed JMVAE (Joint Multimodal VAE), which belongs to the class of approaches that can only learn from the paired  training samples (what we refer to as the \emph{(fully) supervised setting}). It approximates $q(z|x_1, x_2)$, $q_1(z|x_1)$ and $q_2(z|x_2)$ with three corresponding neural networks
and optimizes an ELBO-type loss of the form 
\begin{multline}
 \Lagr = E_{q(z | x_1, x_2)}[\log{p_1(x_1 | z)} + \log{p_2(x_2 | z)}] - \kld{q(z | x_1, x_2)}{\mathcal{N} (0, I)} \\  - \kld{q(z | x_1, x_2)}{q_1(z | x_1)} - \kld{q(z | x_1, x_2)}{q_2(z | x_2)}\enspace.
\end{multline}
The last two  terms imply that during  learning  the joint network output must be generated which requires paired samples. 

\newcommand{\AND}{``AND''}

The MVAE (Multimodal VAE) model \citep{NIPS2018_7801} is the first multimodal VAE-based model allowing for missing modalities that does not require any additional network structures for learning the joint latent code distribution. The joint posterior is modeled using a product-of-experts (PoE, \citealp{hinton2002training,Welling:2007}) as $q(z|x_{1:M}) \propto \prod_{m}q_m(z | x_m)$. For the missing modality $q_k(z | x_k) = 1$ is assumed. 
The model by \cite{NIPS2018_7801}  allows for semi-supervised learning while keeping the number of model parameters low. 
The multiplicative combination of the experts can be interpreted as a conjunction. In the context of probabilistic metric spaces, the product is a triangular norm (t-norm) and generalizes the \AND{} operation to multi-valued logic. 

The bridged model \citep{Yadav2020BridgedVA} highlights the need for an additional network structure for approximating the joint latent code distribution. It attempts to keep the advantages of the additional encoding networks. It  reduces the number of model parameters by introducing the \emph{bridge encoder} that consists of one fully connected layer which takes $z_1$ and $z_2$ latent code vectors generated from $x_1$ and $x_2$ and outputs the mean and the variance of the joint latent code distribution. 

The arguably most advanced multimodal VAE models is VAEVAE by \cite{VAEVAE}, which we discuss in detail in the next section (see also Algorithm~\ref{algorithm1}).

\cite{NIPS2019_9702} proposed a MoE model termed MMVAE (Mixture-of-experts Multimodal VAE). In MMVAE model the joint variational posterior for $M$ modalities is approximated as $q(z|x_{1:M}) = \sum_{m}{\alpha_m}q_m(z | x_m)$ where $\alpha_m = \frac{1}{M}$. The model  utilizes a loss function from the importance weighted autoencoder (IWAE, \citealp{burda2015importance}) that computes a tighter lower bound compared to the VAE ELBO loss. The MoE rule formulation allows in principle to train with a missing modality $i$ by assuming $\alpha_i = 0$, however, \cite{NIPS2019_9702} do not highlight or evaluate this feature. They empirically compare MVAE \citep{NIPS2018_7801} and MMVAE, concluding that MVAE  often fails to learn the joint latent code distribution. Because of these results and those presented by \cite{VAEVAE}, we did not include MVAE as a benchmark model in our experiments.

\section{SVAE}\label{sec:svae}

We developed a new approach as an alternative to VAEVAE. Both models  1) are VAE based; 2) allow for interchangeable cross-model generation as well as a learning joint embedding; 3) allow for missing modalities at training time; and 4) can be applied to two similarly complex high dimensional modalities.
Next, we will present our new model SVAE (it was originally developed to analyse mass spectrometry data, see Section~\ref{sec:spectra}, and thus referred to as SpectraVAE). Then we highlight the differences to VAEVAE. Finally, we state a newly derived objective function for training the models.
We first consider two modalities and generalize to more modalities in Section~\ref{sec:svae-3}.

\subsection{SVAE}
Since both modalities might not be available for all the samples, it should be possible to marginalize each of them
out of $q(z | x_1, x_2)$. While the individual encoding distributions $q(z | x_1)$ and $q(z | x_2)$ can be approximated by neural networks as in the standard VAE, we need to define a meaningful approximation of the joint encoding distribution $q(z | x) = q(z | x_1, x_2)$. 
In the newly proposed SVAE model, these distributions are defined as the following:
\begin{align} 
q(z|x_1, x_2) &= \frac{1}{Z(x_1,x_2)}q_1(z|x_1) q_2(z|x_2)\\ 
q(z|x_1) &= q_1(z | x_1)q^*_2(z | x_1)\label{eq:qzxone}\\ 
q(z|x_2) &= q_2(z | x_2)q^*_1(z | x_2)\label{eq:qzxtwo}\\ 
q(z) &= \mathcal{N} (0, I)
\end{align}
where $Z(x_1,x_2)$ is a normalization constant. The distributions $q_1(z|x_1)$, $q_2(z|x_2)$ and the unormalized  distributions $q^*_2(z | x_1)$ and $q^*_1(z | x_2)$ are approximated by different neural networks. The networks approximating $q_i(z|x_i)$ and $q^*_j(z | x_i)$, $i, j \in \{1, 2\}$ have the same architecture. In case both observations are available, $q(z|x_1, x_2)$ is approximated by applying the product-of-experts rule with $q_1(z|x_1)$ and $q_2(z|x_2)$ being the experts for each modality. In case of a missing modality, equation \ref{eq:qzxone} or \ref{eq:qzxtwo} is used. If, for example, 
 $x_2$ is missing, the $q^*_2(z | x_1)$ distribution takes over as a ``replacement'' expert, 
modelling marginalization over $x_2$.

The model is derived in Section~\ref{sec:svae_aerch}.
The desired properties of the model were that
1) when no modalities are observed the generating distribution for the latent code is Gaussian, 2)
the  modalities are independent given the latent code, 3) both experts cover the whole latent space with equal probabilities, and 4) the joint encoding distribution $q(z | x_1, x_2)$ is modelled by a PoE.

\subsection{Derivation of the SVAE model architecture}\label{sec:svae_aerch}

We define our model in an axiomatic way, requiring the following properties:
\begin{enumerate}
	\item When no modalities are observed, the generating distribution for the latent code is Gaussian:
    \begin{equation}\label{eq1} 
    q(z) = p(z) = \mathcal{N} (0, I)
    \end{equation}
    This property is well known from VAEs and allows easy sampling.
	\item The two modalities are independent  given the latent code, so the decoder distribution is:
	\begin{equation}\label{eqa} 
    p(x_1, x_2|z)=p_1(x_1|z) p_2(x_2|z)
    \end{equation}
    The second property formalizes our goal that the latent representation contains all relevant  information from all modalities.

    The joint  distribution $p(z | x) = p(z | x_1, x_2)$ is given by 
	\begin{multline}\label{eq2new} 
    p(z | x_1, x_2) = \frac{p(z) p_1(x_1,x_2|z)}{p(x_1, x_2) } = \frac{p(z) p_1(x_1,x_2|z)}{\int p(z') p(x_1, x_2|z')  \d z' }\\\overset{(\ref{eqa})}{=} \frac{p(z) p_1(x_1|z) p_2(x_2|z)}{\int p(z') p(x_1|z') p( x_2|z')  \d z' }\enspace.
    \end{multline}
    \item Both experts  cover the whole latent space with equal probabilities:
    \begin{equation}\label{eq3} 
    q(z) = q_1(z)=\int q_1(z|x_1) p(x_1) \d x_1 = \int q_2(z|x_2) p(x_2) \d x_2 = q_2(z) 
    \end{equation}
	\item The joint encoding distribution $q(z | x) = q(z | x_1, x_2)$ is assumed to be given by the product-of-experts rule \citep{hinton2002training,Welling:2007}:
	\begin{equation}\label{eq2} 
    q(z | x_1, x_2) = \frac{1}{Z(x_1, x_2)} q_1(z | x_1)q_2(z | x_2)
    \end{equation}
    with $Z(x_1, x_2) = \int q_1(z' | x_1)q_2(z' | x_2) \d z'$. The modelling by a product-of-experts in \eqref{eq2} is a simplification of \eqref{eq2new} to make the model tractable. 
\end{enumerate}

Given \eqref{eq2} and \eqref{eq3} we obtain
\begin{multline}\label{eq5}
q(z) = \int q(z | x)p(x)\d x = \int q(z | x_1, x_2)p(x_1, x_2)\d x_1 \d x_2  \\ \overset{(\ref{eq2})}{=}   \int \frac{1}{Z(x_1, x_2)}q_1(z | x_1)q_2(z | x_2)p(x_1)p(x_2 | x_1)\d x_1 \d x_2
\enspace.
\end{multline}

Let us define 
\begin{equation}q^*_j(z | x_i) = \int \frac{1}{Z(x_i, x_j)}q_j(z | x_j)p(x_j | x_i)\d x_j
\end{equation}
and write
\begin{multline} \label{eq6}
q(z) \overset{(\ref{eq5})}{=}  \int q_1(z | x_1)p(x_1)\int \frac{1}{Z(x_1, x_2)}q_2(z | x_2)p(x_2 | x_1)\d x_2 \d x_1 \\= \int p(x_1)q_1(z | x_1)q^*_2(z | x_1)\d x_1
\enspace.
\end{multline}

So the proposal distributions are:
\begin{align} \label{eq10}
q(z|x_1, x_2) &= \frac{1}{Z(x_1,x_2)}q_1(z|x_1) q_2(z|x_2)\\ 
q(z|x_1) &= q_1(z | x_1)q^*_2(z | x_1)\\ 
q(z|x_2) &= q_2(z | x_2)q^*_1(z | x_2)\\ 
q(z) &= \mathcal{N} (0, I)
\end{align}

\subsection{SVAE vs.\ VAEVAE}
The VAEVAE model \citep{VAEVAE} is the most similar to ours. 
Wu et al. define two variants which can be derived from the SVAE model in the following way. Variant (a) can be derived by setting  $q^*(z|x_1) = q^*(z|x_2)=1$. Variant (b) is obtained from (a) by additionally using a separate network to model $q(z|x_1,x_2)$.
Having a joint network $q(z|x_1, x_2)$ implements the most straightforward way of capturing the inter-dependencies of the two modalities. However,  the joint network cannot be trained on unpaired data -- which can be relevant when the share of supervised data gets smaller. Option (a) uses the product-of-experts rule to model the joint distribution of the two modalities as well, but does not 
ensure that both experts cover the whole latent space (in contrast to SVAE, see  \eqref{eq3}), which can lead to individual latent code distributions diverging. Based on this consideration and the experimental results from \cite{VAEVAE}, we focused on benchmarking VAEVAE (b) and refer to it as simply VAEVAE in Section \ref{sec:experiments}.

SVAE resembles VAEVAE in the need for additional networks besides one encoder per each modality and the structure of ELBO loss. It does, however, solve the problem of learning the joint embeddings in a way that allows to learn the parameters of approximated $q(z|x_1, x_2)$ using all available samples, i.e., both paired and unpaired. If $q(z|x_1, x_2)$ is approximated with the joint network that accepts concatenated inputs, as in JMVAE and VAEVAE (b), the weights of $q(z|x_1, x_2)$ can only be updated for the paired share of samples. If $q(z|x_1, x_2)$ is approximated with a PoE of decoupled networks as in SVAE, the weights are updated for each sample whether paired or unpaired -- which is the key differentiating feature of SVAE compared to existing architectures.

\subsection{A New Objective Function}
\newcommand{\Lsup}{\Lagr}
\newcommand{\Lcomb}{\Lagr_{\text{comb}}}
When developing SVAE, we devised a novel ELBO-type loss:
\begin{align}
    \Lsup
    = & E_{p_{\text{paired}}(x_1, x_2)}\big[ E_{q(z | x_1, x_2)}[\log{p_1(x_1 | z)}+\log{p_2(x_2 | z)}]\notag\\
    &- \kld{q(z | x_1, x_2)}{p(z | x_1)} - \kld{q(z | x_1, x_2)}{p(z | x_2)}\big]\notag\\
    &+ E_{p_{\text{paired}}(x_1)}\left[ E_{q(z | x_1)}[\log{p_1(x_1 | z)}] - \kld{q(z | x_1)}{p(z)}\right] \notag\\
    &+ E_{p_{\text{paired}}(x_2)}\left[ E_{q(z | x_2)}[\log{p_2(x_2 | z)}] - \kld{q(z | x_2)}{p(z)}\right]\label{eq:suploss}\\
    \Lagr_{1}
    =& E_{p_{\text{unpaired}}(x_1)}\left[ E_{q(z | x_1)}[\log{p_1(x_1 | z)}] - \kld{q(z | x_1)}{p(z)}\right]\\
    \Lagr_{2}
    =& E_{p_{\text{unpaired}}(x_2)}\left[ E_{q(z | x_2)}[\log{p_2(x_2 | z)}] - \kld{q(z | x_2)}{p(z)}\right]\\
    \Lcomb =& \Lsup + \Lagr_{1} + \Lagr_{2}\label{loss_function}
\end{align}
Here $p_{\text{paired}}$ and $p_{\text{unpaired}}$ denote the distributions of the paired and unpaired training data, respectively.
The differences between this loss and the loss function used to train VAEVAE by \cite{VAEVAE} are highlighted in Algorithm \ref{algorithm1}.

The  loss function can derived as follows. 
Let consider the optimization problem
\begin{multline}\label{eq_sum_of_p}
    E_{p_{\text{Data}}(x_1,x_2)}[\log p(x_1,x_2)] = \\ = \frac 1 2 E_{p_{\text{Data}}(x_1,x_2)}[\log p(x_1|x_2) + \log p(x_2) + \log p(x_2|x_1) + \log p(x_1)]\\
    = \frac 1 2 E_{p_{\text{Data}}(x_1,x_2)}[\log p(x_1|x_2)]
    + \frac 1 2 E_{p_{\text{Data}}(x_1,x_2)}[\log p(x_2|x_1)]\\
    + \frac 1 2 E_{p_{\text{Data}}(x_2)}[\log p(x_2)]
    + \frac 1 2 E_{p_{\text{Data}}(x_1)}[\log p(x_1)]\enspace.
\end{multline}
We can now proceed by finding lower-bounds for each term. 
For the last two terms $\log p(x_i)$ we can use the standard ELBO as given in \eqref{eq_ELBO_single}. This gives the terms 
\begin{equation}\label{eq_ELBO_single_2}
    \Lagr_i = E_{p_{\text{Data}}(x_i)}\left[ E_{q(z | x_i)}[\log{p_i(x_i | z)}] - \kld{q(z | x_i)}{p(z)}\right]
\end{equation}

Next, we will derive $\log p(x_1|x_2)$. This we can do in terms of a conditional VAE \citep{NIPS2015_5775}, where we condition all terms on $x_2$ (or $x_1$ if we model $\log p(x_2|x_1)$). So  we derive the log-likelihood for 
$p(x_1|x_2) = \int p(x_1| z) p(z|x_2) dz$, where $p(z|x_2)$ is now our prior.
By model assumption we further have $p(x_1,x_2,z)= p(x_1|z) p(x_2|z) p(z)$ and therefore $p(x_1|x_2, z) =  p(x_1|z)$. Thus we arrive at the ELBO losses
\begin{equation}
    \Lagr_{12} = E_{p_{\text{Data}}(x_1, x_2)}\left[ E_{q(z | x_1, x_2)}[\log{p_1(x_1 | z)}] - \kld{q(z | x_1, x_2)}{p(z | x_2)}\right]
\end{equation}
and
\begin{equation}
    \Lagr_{21} = E_{p_{\text{Data}}(x_1, x_2)}\left[ E_{q(z | x_1, x_2)}[\log{p_2(x_2 | z)}] - \kld{q(z | x_1, x_2)}{p(z | x_1)}\right]\enspace.
\end{equation}
We now insert the terms in \eqref{eq_sum_of_p} and arrive at:
\begin{multline}\label{eq_sum_of_p_multi}
    2 E_{p_{\text{Data}}(x_1,x_2)}[\log p(x_1,x_2)] \geq  \Lagr_{12}+\Lagr_{21} + \Lagr_{1} + \Lagr_{2} \\
    = E_{p_{\text{Data}}(x_1, x_2)}\left[ E_{q(z | x_1, x_2)}[\log{p_1(x_1 | z)}] - \kld{q(z | x_1, x_2)}{p(z | x_2)}\right]\\
    + E_{p_{\text{Data}}(x_1, x_2)}\left[ E_{q(z | x_1, x_2)}[\log{p_2(x_2 | z)}] - \kld{q(z | x_1, x_2)}{p(z | x_1)}\right]\\
    + E_{p_{\text{Data}}(x_1)}\left[ E_{q(z | x_1)}[\log{p_1(x_1 | z)}] - \kld{q(z | x_1)}{p(z)}\right]\\
    + E_{p_{\text{Data}}(x_2)}\left[ E_{q(z | x_2)}[\log{p_2(x_2 | z)}] - \kld{q(z | x_2)}{p(z)}\right]\enspace.
\end{multline}

The first two terms together give 
\begin{equation}
    E_{p_{\text{Data}}(x_1, x_2)}\left[ E_{q(z | x_1, x_2)}[\log{p_1(x_1 | z)}+\log{p_2(x_2 | z)}]\right]\enspace.
\end{equation}

We do not know the conditional prior $p(z | x_i)$. By definition of the VAE, we are allowed to optimize the prior, therefore we can parameterize it and optimize it. However, we know that in an optimal model $p(z | x_i)\approx q(z | x_i)$ and it might be possible to prove that if $p(z | x_i)$ is learnt in the same model-class as $q(z | x_i)$ we can find that the optimum is indeed $p(z | x_i)=q(z | x_i)$. Inserting this choice into the equation gives the end-result.

\renewcommand{\KL}{D_{\text{KL}}}
\begin{algorithm}[H]
\SetAlgoLined
\SetKwInOut{Input}{Input}
\Input{Supervised example $(x_1, x_2)$, unsupervised example $x_1'$, unsupervised example $x_2'$}
$z' \sim q(z | x_1, x_2)$\\
$z_{x_1} \sim q_1(z | x_1)$\\
$z_{x_2} \sim q_2(z | x_2)$\\
$d_1 = \KL(q(z' | x_1, x_2) \| q_1(z_{x_1} | x_1)) \boldsymbol{+ \KL(q_1(z_{x_1} | x_1) \|p(z))}$\\
$d_2 = \KL(q(z' | x_1, x_2) \| q_2(z_{x_2} | x_2)) \boldsymbol{+ \KL(q_2(z_{x_2} | x_2) \| p(z))}$\\
$\Lsup = \log{p_1(x_1 | z)} + \log{p_2(x_2 | z)} + \boldsymbol{\log{p_1(x_1 | z_{x_1})}} + \boldsymbol{\log{p_2(x_2 | z_{x_2})}} + d_1 + d_2$\\
$\Lagr_{x_1} = \log{p_1(x_1' | z_{x_1})} + \KL(q_1(z_{x_1} | x_1') \| p(z))$\\
$\Lagr_{x_2} = \log{p_2(x_2' | z_{x_2})} + \KL(q_2(z_{x_2} | x_2') \| p(z))$\\
$\Lcomb = \Lagr + \Lagr_{x_1} + \Lagr_{x_2}$
 \caption{Loss computation (forward pass) for SVAE and VAEVAE*. In bold are terms that are different from \cite{VAEVAE}}
 \label{algorithm1}
\end{algorithm}

\newcommand{\Lvaevae}{\Lagr_{\text{comb}}}
\subsection{SVAE and VAEVAE for more than two modalities}\label{sec:svae-3}
In the following, we 
formalize the VAEVAE model for three modalities and present a na\"ive 
extension of the SVAE model to more than two modalities.
\begin{figure}[ht!]
  \centering
  \includegraphics[width=1\linewidth]{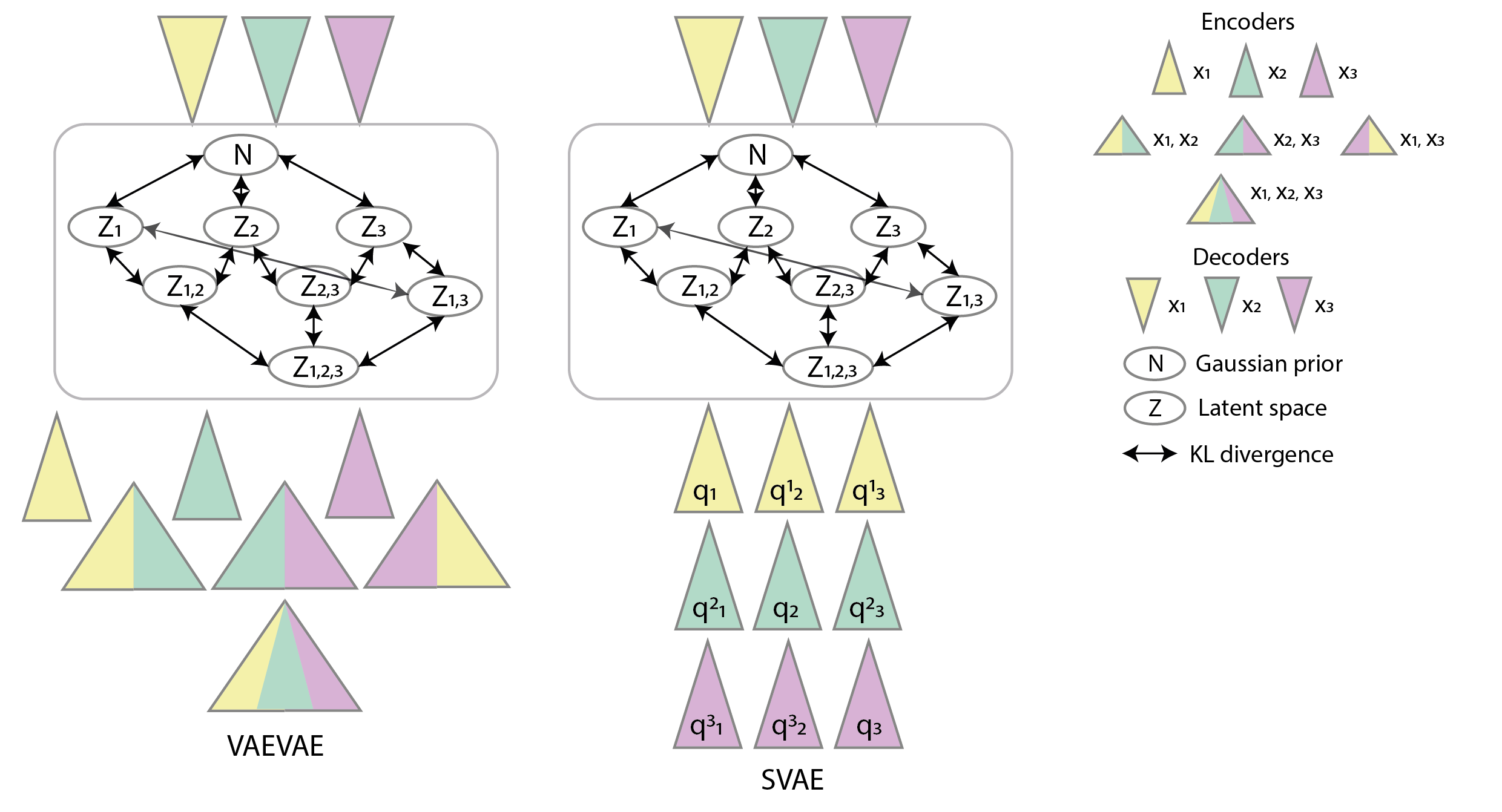}
  \caption{The SVAE and VAEVAE network architectures for 3-modalities. The number of parameters is $kn^2$ for SVAE and $kn2^{n-1}$ for VAEVAE, where $n$ is the number of modalities and $k$ is the number of parameters in one encoding network.\label{fig:architecture3}}
\end{figure}

In the canonical extension of 
VAEVAE to three modalities,  the three- and two-modal relations are captured by the corresponding networks $q(z|x_1, x_2, x_3)$, $q(z|x_i, x_j)$ and $q(z|x_i)$ for $i,j\in\{1,2,3\}$, see Figure~\ref{fig:architecture3}. In the general $n$-modal case, the model has $2^n$ networks. For $n=3$, the loss function reads:

\begin{align}
    \Lsup_{1,2,3}
    = & E_{p_{\text{paired}}(x_1, x_2, x_3)}\left[ E_{q(z | x_1, x_2, x_3)}[\log{p_1(x_1 | z)}+\log{p_2(x_2 | z)}+\log{p_3(x_3 | z)}]\right]\notag\\
    &- \kld{q(z | x_1, x_2, x_3)}{q(z | x_1,x_2)}\notag\\
    &- \kld{q(z | x_1, x_2, x_3)}{q(z | x_2,x_3)}\notag\\
    &- \kld{q(z | x_1, x_2, x_3)}{q(z | x_1,x_3)}\label{loss_function_three_one}\\
    \Lagr_{ij}
    = & E_{p_{\text{paired}}(x_1, x_2, x_3)}\left[ E_{q(z | x_i, x_j)}[\log{p_i(x_i | z)}+\log{p_j(x_j | z)}]\right]\notag\\
    &- \kld{q(z | x_i, x_j)}{q(z | x_1)} - \kld{q(z | x_i, x_j)}{q(z | x_2)}\notag\\
    &- \kld{q(z | x_i, x_j)}{q(z | x_3)} - \kld{q(z | x_i, x_j)}{q(z)}\\
    \Lagr_{i}
    =& E_{p_{\text{unpaired}}(x_i)}\left[ E_{q(z | x_i)}[\log{p_i(x_i | z)}] - \kld{q(z | x_i)}{q(z)}\right]\\
    \Lvaevae =& \Lsup_{1,2,3} + \sum\limits_{i,j\in\{1,2,3\},i \neq j} \Lagr_{i,j} + \sum\limits_{i=1}^3 \Lagr_{i}\label{loss_function_three_four}
\end{align}

In this study, we considered a simplifying extension of SVAE to $n$ modalities using  $n^2$ networks $q_i^j(z|x_j)$ for $i,j\in\{1,\dots,n\}$. For the 3-modal case depicted in Figure~\ref{fig:architecture3}, the PoE relations between the modalities are defined in the following way:
\begin{align} 
q(z|x_1, x_2, x_3) &= \frac{1}{Z(x_1,x_2,x_3)}q_1^1(z|x_1) q_2^2(z|x_2) q_3^3(z|x_3)\\ 
i,j,k\in\{1,2,3\},i \neq j \neq k:\\
q^i(z|x_i, x_j) &= \frac{i}{Z(x_i,x_j)}q_i^i(z|x_i) q_j^j(z|x_j) q_k^i(z|x_i)\\ 
q^j(z|x_i, x_j) &= \frac{1}{Z(x_i,x_j)}q_i^i(z|x_i) q_j^j(z|x_j) q_k^j(z|x_j)\\ 
q(z|x_i) &= q_i^i(z|x_i) q_j^i(z|x_i) q_k^i(z|x_i)\\ 
q(z) &= \mathcal{N} (0, I)
\end{align}
The corresponding SVAE loss function has additional terms due to the fact that the relations between pairs of modalities need to be captured with two PoE rules $q^i(z|x_i, x_j)$ and $q^i(z|x_i, x_j)$ in SVAE, while there is  only a single network $q(z|x_i, x_j)$ in VAEVAE. The  loss functions \eqref{loss_function_three_one}--\eqref{loss_function_three_four} above are modified in a way that $f(q(z|x_i, x_j)) = f(q^i(z|x_i, x_j)) + f(q^j(z|x_i, x_j))$ for any function $f$.

This  extension of the bi-modal case assumes that $p(x_i, x_j|x_k) = p(x_i|x_k) p(x_j|x_k)$ for $i,j,k\in\{1,2,3\},i \neq j \neq k$, which implies that $x_i$, $x_j$ and $x_k$ are independent of each other.

\section{Experiments}\label{sec:experiments}
\begin{figure}[t!]

\subfloat{%
  \includegraphics[width=0.9\linewidth]{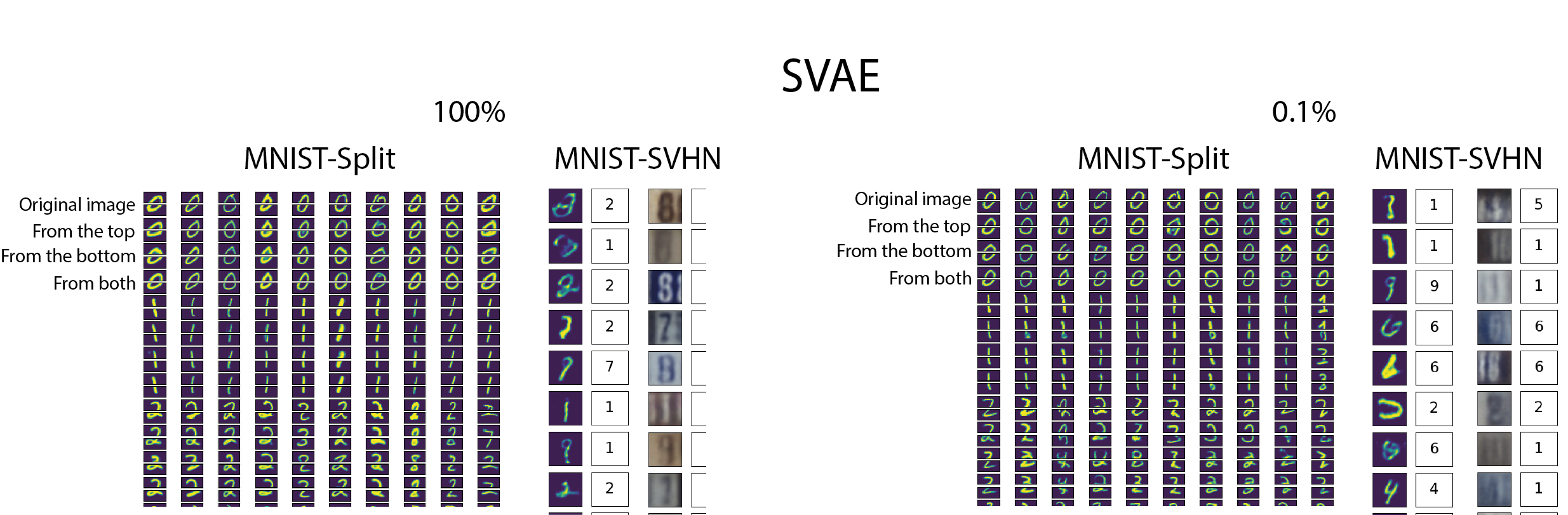}
}

\subfloat{%
  \includegraphics[width=0.9\linewidth]{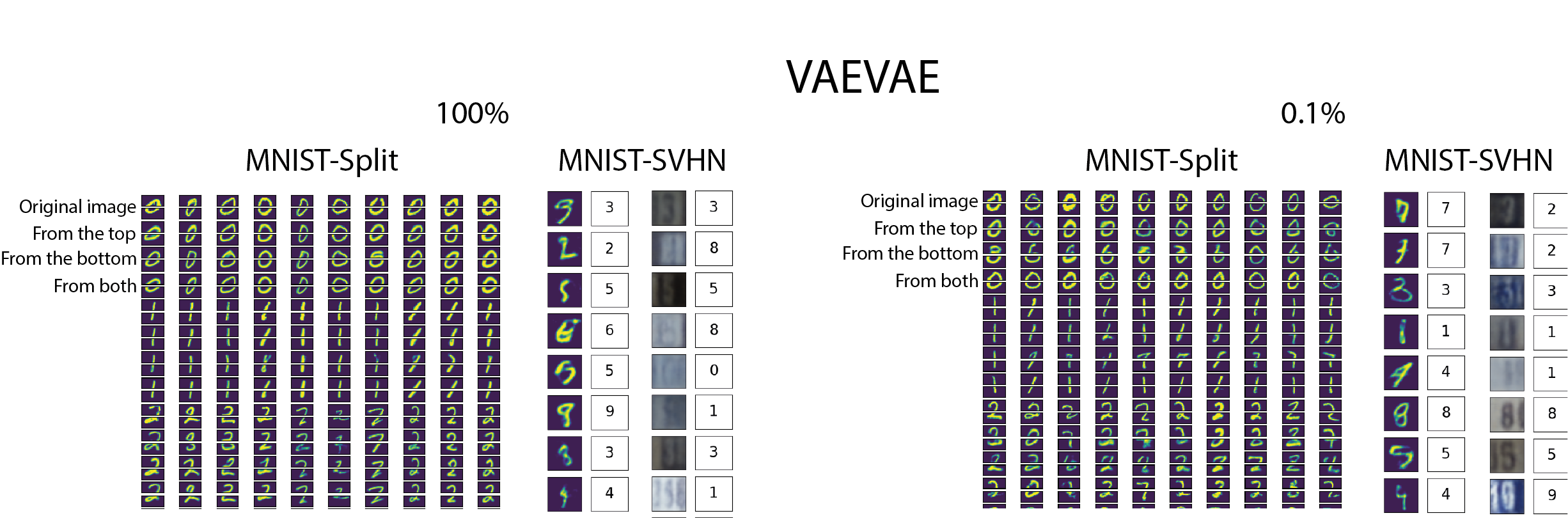}
}

\caption{MNIST-Split image reconstructions of a top half and a bottom half given the top half, the bottom half of the original image or both halves. Side-by-side MNIST-SVHN reconstruction from randomly sampled latent space, with oracle predictions of a digit class. The joint coherence is a share of classes predicted the same. The examples are generated by SVAE and VAEVAE for the supervision levels 100\% and 0.1\%}
\label{fig:qual_svae}
\end{figure}

We conducted experiments to compare state-of-the-art PoE based VAEs 
with the MoE approach MMVAE \citep{NIPS2019_9702}.
We considered  VAEVAE (b) as proposed by \cite{VAEVAE} and SVAE as described above. The two approaches differ both in the underlying model as well as the objective function. For a better understanding of these differences, we also considered an algorithm referred to as VAEVAE*, which  has the same model architecture as VAEVAE and the same loss function as SVAE.\footnote{We also evaluated SVAE*, our model with the VAEVAE loss function, but it never outperformed other models.} The difference in the training procedure for VAEVAE and VAEVAE* is shown in Algorithm~\ref{algorithm1}. Since the VAEVAE implementation was not publicly available at the time of writing, we used our own implementation of VAEVAE based on the PiXYZ library.\footnote{https://github.com/masa-su/pixyz} For details about the experiments we refer to Appendix~\ref{app:training}. The source code to reproduce the experiments can be found at \url{https://github.com/sgalkina/poe-vaes}. More qualitative examples are shown in Figure \ref{fig:qual_svae} for SVAE and VAEVAE.

For an unbiased evaluation, we considered the same test problems and performance metrics as \cite{NIPS2019_9702}. In addition, we designed an experiment referred to as MNIST-Split that was supposed to be well-suited for PoE.
In all experiments we kept the network architectures as similar as possible (see Appendix~\ref{app:training}).
For the new benchmark problem, we  constructed a multi-modal dataset where  the  modalities are similar in dimensionality as well as  complexity and are providing missing information to each other rather than duplicating it. 
The latter should favor a PoE modelling, which suits an ``AND'' combination of the modalities.

We measured  performance for different supervision levels for each dataset (e.g., 10\% supervision level means that 10\% of the training set samples were  paired and 
the remaining 90\% were unpaired).

Finally, we compared VAEVAE and SVAE
on the real-world bioinformatics task that motivated our study, namely learning a joint  
representation of mass spectra and molecule structures.

\subsection{Image and image: MNIST-Split}
\begin{figure}[t!]
  \centering
  \includegraphics[width=0.8\linewidth]{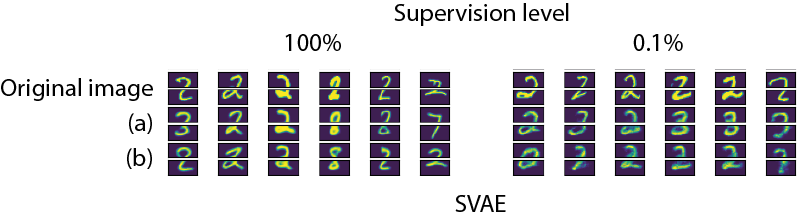}
  \caption{MNIST-Split image reconstructions of a top half and a bottom half given (a) the top half; (b) the bottom half of the original image.}
  \label{fig:qual_mnist_split}
\end{figure}
We created an image reconstruction dataset based on MNIST digits \citep{mnist}. 
The images were split horizontally into equal parts, either two or three depending on the experimental setting. These regions are considered as different input modalities.

Intuitively, for this task a joint representation should encode the latent class labels, the digits.  
The correct digit can sometimes be inferred from only one part of the image (i.e., one modality), but sometimes both modalities are needed. 
In the latter cases, an \AND{} combination of the inputs is helpful.
This is in contrast to the MNIST-SVHN task described below, where the joint label could in principle be inferred from each input modality independently.

\begin{figure}
  \begin{center}
    \includegraphics[width=0.25\textwidth]{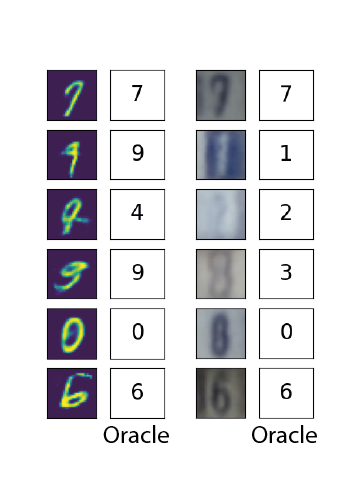}
  \end{center}
  \caption{MNIST-SVHN reconstruction for fully supervised VAEVAE.}
  \label{fig:qual_svhn}
\end{figure}

\paragraph{Two modalities: MNIST-Split.}
In the bi-modal version referred to as MNIST-Split, the MNIST images were split in top and bottom halves of equal size, and the halves were then used as two modalities.
We tested the quality of the image reconstruction given one or both modalities by predicting the reconstructed image label with an independent oracle network,  a  ResNet-18 \citep{7780459} trained on the original MNIST dataset. The evaluation metrics were \emph{joint coherence}, \emph{synergy}, and \emph{cross-coherence}.
For measuring joint coherence, 1000 latent space vectors were generated from the prior and both halves of an image were then reconstructed with the corresponding decoding networks. The concatenated halves yield the fully reconstructed image. Since the ground truth class labels do not exist for the randomly sampled latent vectors, we could only perform  a qualitative evaluation, see Figure \ref{fig:qual_mnist_split}.
Synergy was defined as the accuracy of the image reconstruction given both halves.
{Cross-coherence} considered the reconstruction of the full image from one half and was defined as the fraction of class labels correctly predicted by the oracle network.

\begin{table}[th!]
\caption{Evaluation of the models trained on the fully supervised datasets.\label{tab:eqsplitres}}
\begin{center}
\begin{tabular}{@{} l r r r r @{}}
\toprule
& Accuracy (both) & Accuracy (top half) & Accuracy (bottom half) \\ 
 \midrule
MMVAE & 0.539 & 0.221 & 0.283 \\ 
SVAE & 0.948 & 0.872 & 0.816 \\ 
VAEVAE & 0.956 & 0.887 & 0.830 \\  
VAEVAE* & 0.958 & 0.863 & 0.778 \\ 
\bottomrule
\end{tabular}
\end{center}

\end{table}

\begin{figure}[ht]
  \centering
\subfloat{%
  \includegraphics[clip,width=\columnwidth]{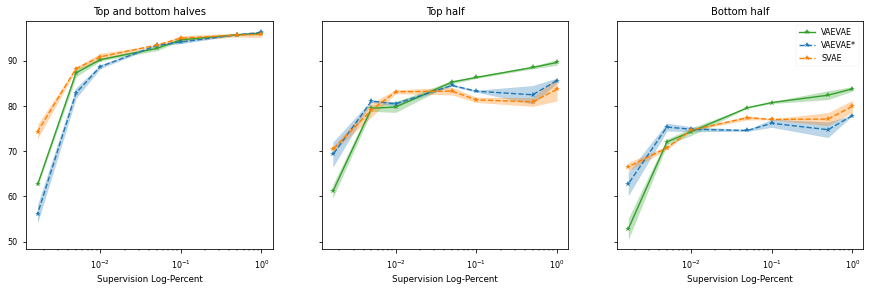}%
}
  \caption{MNIST-Split dataset. Accuracy of an oracle network applied to images  reconstructed given the full image (both halves), the top half and the bottom half.\label{fig:images_images}}
\end{figure}

The quantitative results are shown in Table~\ref{tab:eqsplitres} and Figure~\ref{fig:images_images}. All  PoE  architectures clearly outperformed MMVAE even when trained on the low supervision levels.
In this experiment, it is important that both experts agree on a class label. Thus, as expected, the multiplicative PoE fits the task much better than the additive mixture.
Utilizing the novel loss function  (\ref{loss_function}) gave the best results for very low supervision (SVAE and VAEVAE*). 

\begin{figure}[ht!]
  \centering
  \includegraphics[width=1\linewidth]{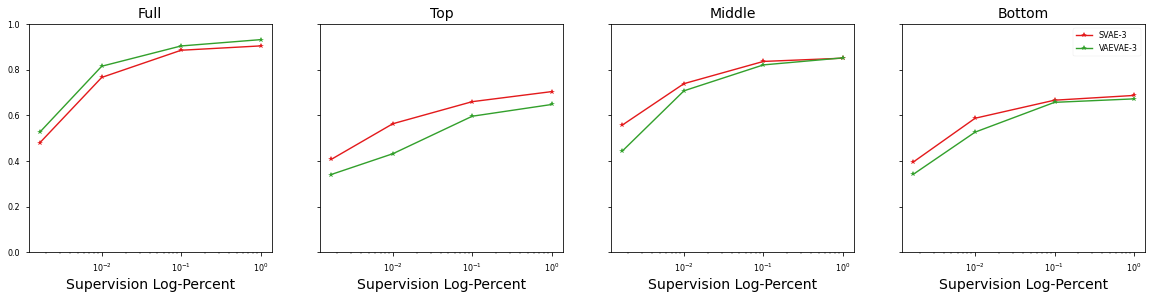}
  \caption{MNIST-Split-3 dataset, reproducing the logic of MNIST-Split for the input images split into three parts. The plots show the accuracy of an oracle network getting the full image or a single of  the three single modalities (Top, Middle, Button) as inputs. \label{fig:images_images3}}
\end{figure}

\paragraph{Three modalities: MNIST-Split-3.}
We compared a simple generalization of the SVAE model to more than two modalities with the canonical extension of the VAEVAE model described in Section~\ref{sec:svae-3} on the MNIST-Split-3 data, the 3-modal version of MNIST-Split task.
Figure~\ref{fig:images_images3} shows that SVAE performed better when looking at the reconstructions from individual modalities, but worse when all three modalities are given.
While the number of parameters in the bi-modal case is the same for SVAE and VAEVAE, it grows exponentially for VAEVAE and stays in order of $n^2$ for SVAE where $n$ is the number of modalities, see Figure~\ref{fig:architecture3} and Section~\ref{sec:svae-3} for details. 

\subsection{Image and image: MNIST-SVHN}
\begin{figure}
  \centering
  \includegraphics[width=1\linewidth]{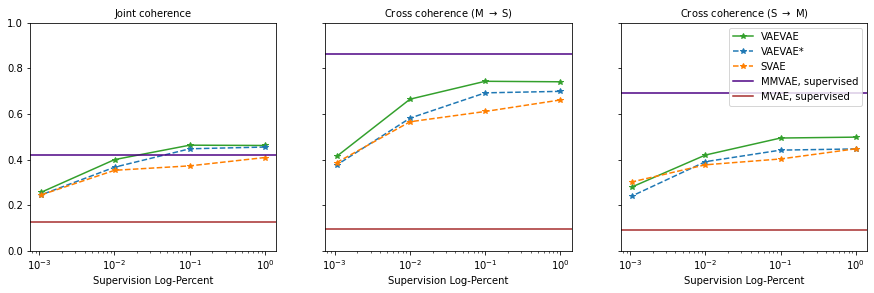}
  \caption{Performance on MNIST-SVHN for different supervision levels. (left) Joint coherence, a share of generated images with the same digit class; (middle) Cross-coherence, accuracy of SVHN reconstructions given MNIST; (right) Cross-coherence, accuracy of MNIST reconstructions given SVHN.}
  \label{fig:sup_coherenc_ms}
\end{figure}
The first dataset considered by \cite{NIPS2019_9702} is constructed by pairing MNIST and SVHN \citep{37648} images showing the same digit. This dataset shares some properties with MNIST-Split, but the relation between the two modalities is different: the digit class is derived from a concatenation of two modalities in MNIST-Split, while in MNIST-SVHN it could be derived from any modality alone. As before,  oracle networks are trained to predict the digit classes of MNIST and SVHN images. Joint coherence was again computed based on 1000 latent space vectors  generated from the prior. Both images were then reconstructed with the corresponding decoding networks. A reconstruction was considered correct if the predicted digit classes of MNIST and SVHN were the same.
Cross-coherence was measured as above. 

Figure \ref{fig:qual_svhn} shows examples of paired image reconstructions from the randomly sampled latent space of the fully supervised VAEVAE model. The digit next to the each reconstruction shows the digit class prediction for this image. 
The quantitative results in  Figure \ref{fig:sup_coherenc_ms} show that
all three PoE based models reached a similar joint coherence as MMVAE, VAEVAE scored even higher. The cross-coherence results were best for MMVAE, but the three PoE based models performed considerably better  than the MVAE baseline reported by \cite{NIPS2019_9702}.

\subsection{Image and text: CUB-Captions}
The second benchmark considered by  \cite{NIPS2019_9702} is the CUB Images-Captions dataset \citep{WahCUB_200_2011} containing photos of birds and their textual descriptions. Here the  modalities are of different  nature but similar in dimensionality and information content.
We used the source code\footnote{https://github.com/iffsid/mmvae} by Shi et al. to compute the same evaluation metrics as in the MMVAE study. Canonical correlation analysis (CCA) was used for estimating joint and cross-coherences of images and text  \citep{massiceti2018visual}. The projection matrices $W_x$ for images and $W_y$ for captions were pre-computed using the training set of CUB Images-Captions and are available as part of the source code. Given a new image-caption pair $\tilde{x}, \tilde{y}$, we computed the correlation between the two by
$\operatorname{corr}(\tilde{x}, \tilde{y}) = \frac{\phi(\tilde{x})^T\phi(\tilde{y})}{\norm{\phi(\tilde{x})}\norm{\phi(\tilde{y})}}
$, where
 $\phi(\tilde{k}) = W_k^T\tilde{k} - \operatorname{avg}(W_k^Tk)$.

We employed the same image generation procedure as in the MMVAE study. Instead of creating the images directly, we generated  2048-d feature vectors using a pre-trained ResNet-101. In order to find the resulting image, a nearest neighbours lookup with Euclidean distance was performed. A CNN encoder and decoder was used for the  (see Table~\ref{tab:cupcap_a} and Table~\ref{tab:cupcap_b}). Prior to computing the correlations, the captions were converted to 300-d vectors using FastText \citep{Bojanowski_2017}. As in the experiment before, we used the same network architectures and hyperparameters as \cite{NIPS2019_9702}.   We sampled  1000 latent space vectors  from the prior distribution. Images and captions were then reconstructed with the decoding networks. The joint coherence was then computed as the CCA  for the resulting image and caption averaged over the 1000 samples.
{Cross-coherence} was computed from caption to  image and vice versa using the CCA  averaged over the whole test set.

\begin{figure}[ht!]
  \centering
  \includegraphics[width=1\linewidth]{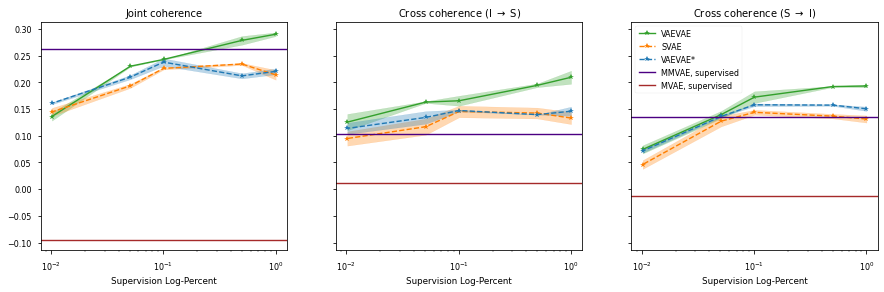}
  \caption{CUB Images-Captions dataset. Performance metrics for different supervision levels. (left) Joint coherence, the correlation between images and labels reconstructed from the randomly sampled latent vectors; (middle) Cross-coherence, the correlation of the reconstructed caption given the image; (right) Cross-coherence, the correlation of the reconstructed image given the caption.}
  \label{fig:sup_coherence}
\end{figure}

\begin{figure}[ht!]
  \centering
  \includegraphics[width=1\linewidth]{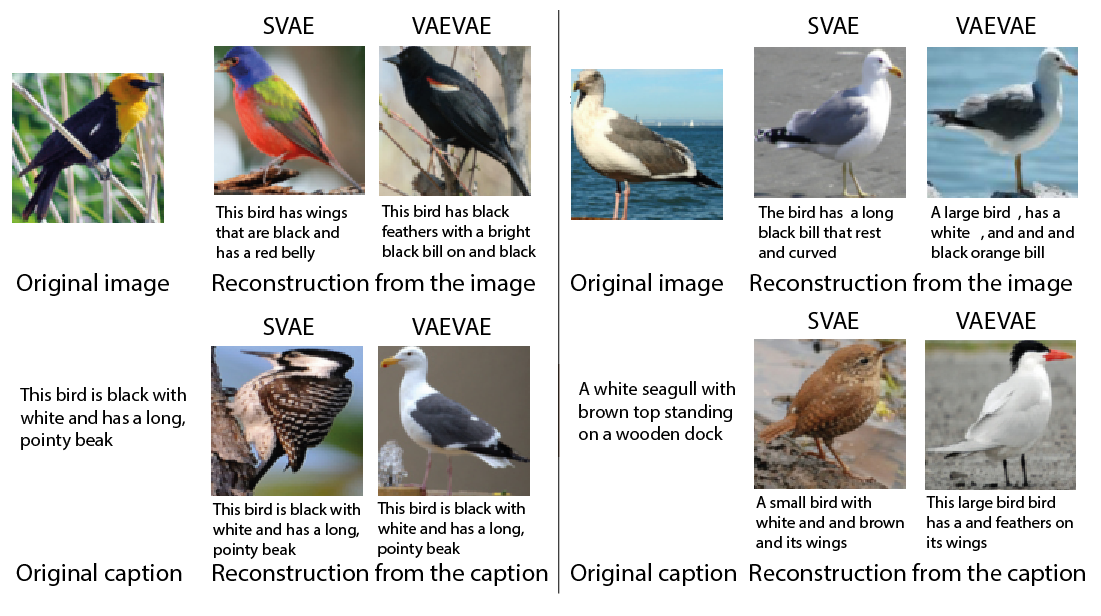}
  \caption{Examples of image and caption reconstructions given one modality input for SVAE and VAEVAE. Given that the caption can be broad (e.g., "this bird is black and white and has a long pointy beak" in the example), it can fit many different images. In this case, the image from the caption reconstruction tends to better fit the description than the original image. The same goes for images: one of the reconstructed images has a bird with a red belly which got reflected in the generated caption even though it was not a part of the original caption.}
\end{figure}

As can be seen in Figure~\ref{fig:sup_coherence}, VAEVAE showed the best performance among all models.
With  full supervision the VAEVAE model outperformed MMVAE in all three metrics. The cross-coherence of the three PoE models was higher or equal to MMVAE except for very low supervision levels. All three PoE based models were consistently better than MVAE. 

\subsection{Chemical structures and mass spectra}
\label{sec:spectra}

We evaluated SVAE and VAEVAE on a real-world bioinformatics application. The models were used for annotating mass spectra with molecule structures. Discovering chemical composition of a biological sample is one of the key problems in analytical chemistry. Mass spectrometry is a common high throughput analytical method. Thousands of spectra can be generated in a short time, but identification rates of the corresponding molecules is still low for most of the studies. 

We approach the spectra annotation problem with bi-modal VAEs in the following way: the SVAE and VAEVAE models are trained on a subset of the MoNA (Mass Bank of North America) dataset \citep{Vinaixa2016}, where the mass spectra suitable for molecule identification are assembled. We focused on tandem mass spectra from only one type of mass spectrometer collected in the positive ion mode. The mass spectra were the first modality. To represent a molecule structure, we used a molecule structural fingerprint, a bit string where an individual bit shows if a substructure from a predefined set of possible substructures is present in the molecule. Fingerprints of length $2149$ were used as the second modality. During testing, only the mass spectrum was provided and the fingerprint was predicted. A molecule structure still has to be identified based on the predicted structural fingerprint. We did this by ranking the candidate molecules from a molecule database based on the cross entropy loss between the molecule fingerprint and the predicted fingerprint.

\begin{figure}[ht!]
  \centering
  \includegraphics[width=1\linewidth]{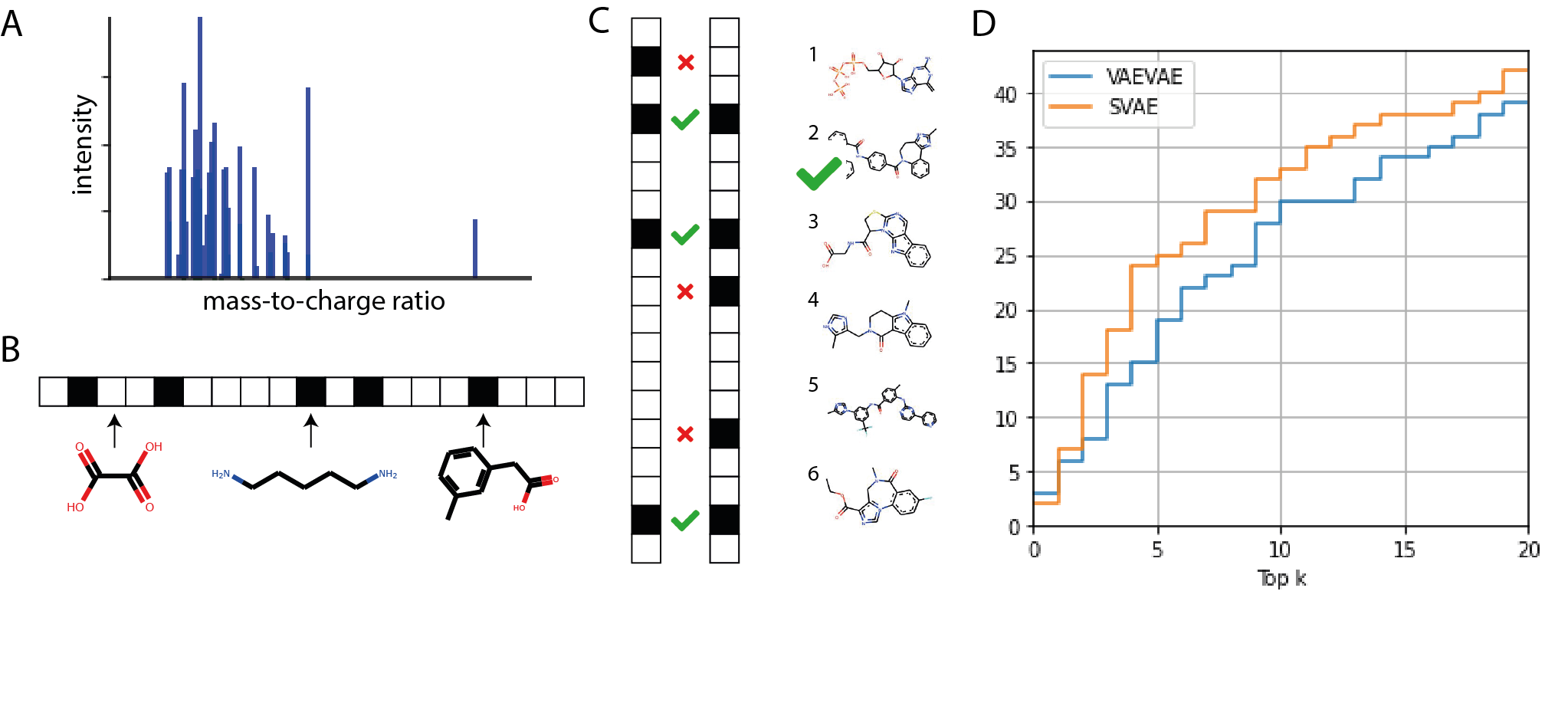}
  \caption{A) The first modality: a tandem mass spectrum; B) The second modality: a molecule structure fingerprint, a bit string where each bit represents if a given substructure is present in the molecule; C) The performance evaluation: the fingerprint is predicted with only the spectrum as  input. The candidate molecules are ranked by cross-entropy loss between their fingerprints and the predicted fingerprint. The rank of the correct candidate is used for comparing the performance of different methods. D) The evaluation results for CASMI2017 challenge, with 112 test spectra and 200-10000 candidate molecules for each spectrum: The plot shows for how many spectra the correct candidate appeared in top $k$ candidates. SVAE performs better than VAEVAE in this preliminary evaluation.\label{fig:molecule}}
\end{figure}

For evaluation, we used competition data from the CASMI2017 (Critical Assessment of Small Molecule Identification) challenge \citep{Schymanski2017}. Since the molecule structure identification is based on ranking the candidate list, we focused on the part of the challenge where the candidate list is already provided for each spectrum. In the preliminary experiments, SVAE outperformed VAEVAE in predicting molecule structural fingerprints from mass spectra, see Figure~\ref{fig:molecule}.

\section{Discussion and Conclusions}
We studied bi-modal variational autoencoders (VAEs) based on a product-of-experts (PoE)architecture, in particular VAEVAE as proposed by \cite{VAEVAE} and a new model SVAE, which we derived in an axiomatic way, and represents a generalization of the VAEVAE architecture. The models learn representations that allow coherent sampling of the modalities and accurate sampling of one modality given the other. They work well in the semi-supervised setting, that is, not all modalities need to be always observed during training.
It has been argued that the mixture-of-experts (MoE) approach MMVAE is preferable to a PoE  for multimodal VAEs \citep{NIPS2019_9702}, in particular in the fully supervised setting (i.e., when all data are paired). This conjecture was based on a comparison with the MVAE model \citep{NIPS2018_7801}, but is refuted by our experiments showing that VAEVAE  and our newly proposed SVAE can outperform MMVAE on experiments conducted by \cite{NIPS2019_9702}. 
Intuitively,  
PoEs are more tailored to towards an ``AND'' (multiplicative) combination of the input modalities. This is supported by our experiments on halved  digit images, where a conjunctive combination is helpful and the PoE models perform much better than MMVAE. 
In a real-world bioinformatics task,
 SVAE outperformed VAEVAE in predicting molecule structural fingerprints from mass spectra.
We also expanded SVAE and VAEVAE to 3-modal case and show that SVAE demonstrates better performance on individual modalities reconstructions while having less parameters than VAEVAE.

 \bibliographystyle{elsarticle-num} 
 \bibliography{cas-refs}

\renewcommand\thefigure{\thesection.\arabic{figure}} 
\renewcommand\thetable{\thesection.\arabic{table}} 
\renewcommand{\theequation}{\thesection.\arabic{equation}}

\appendix

\section{Details of experiments}
\label{app:training}

The encoder and decoder architectures for each experiment and modality are listed below. To implement joint encoding network (VAEVAE architecture), an fully connected layer followed by ReLU is added to the encoding architecture for each modality. Another fully connected layer accepts the concatenated features from the two modalities as an input and outputs the latent space parameters. Adam optimiser is used for learning in all the models \citep{kingma2014method}. We used a padding of 1 pixel if the stride was 2 pixels and no padding otherwise.

\paragraph{MNIST-Split.}  

The models are trained for 200 epochs with the learning rate $2\cdot10^{-4}$. The best epoch is chosen by the highest accuracy of the reconstruction from the top half evaluated on the validation set.
We used a latent space dimensionality of $L=64$.  The network architectures are described in Table~\ref{tab:split_arch}.

\begin{table}[hbt!]
\begin{center}
\begin{minipage}{.49\textwidth}
\begin{center}
\begin{tabular}{@{} l @{}}
\toprule
\textbf{Encoder}\\ 
\midrule
Input $\in \mathbb{R}^{3\times32\times32}$ \\
 $4\times4$ conv. 64 stride 2 ReLU \\
 $4\times4$ conv. 128 stride 2 ReLU \\
 $4\times4$ conv. 256 stride 2 ReLU \\
FC. 786  ReLU \\
FC. L, FC. L \\
\bottomrule
\end{tabular}
\end{center}
\end{minipage}\begin{minipage}{.49\textwidth}
\begin{center}
\begin{tabular}{@{} l  @{}}
\toprule
\textbf{Decoder}\\ 
\midrule
Input $\in \mathbb{R}^{L}$ \\
FC. L  ReLU \\
FC. 512  ReLU \\
FC. 112  ReLU \\
 $4\times4$ upconv. 56 stride 1 ReLU \\
 $4\times4$ upconv. 28 stride 2 ReLU \\
\bottomrule
\end{tabular}
\end{center}
\end{minipage}
\end{center}
\caption{Network architectures for MNIST-Split for each image half.\label{tab:split_arch}}
\end{table}

\paragraph{MNIST-SVHN.}  

The models were trained for 50 epochs with  learning rates $10^{-3}$ and $10^{-4}$. Only the results for the best learning rate are reported ($10^{-3}$ for VAEVAE and VAEVAE* and $10^{-4}$ for SVAE). The best epoch was chosen based on the highest joint coherence evaluated on the validation set.
We used a latent space dimensionality of $L=20$. The network architectures are summarized in Table~\ref{tab:mnist_arch} and Table~\ref{tab:svhn_arch} for the MNIST and SVHN modality, respectively.

\begin{table}[hbt!]
\begin{center}
\begin{minipage}{.3\textwidth}
\begin{center}
\begin{tabular}{@{} l  @{}}
\toprule
\textbf{Encoder}\\ 
\midrule
Input $\in \mathbb{R}^{1\times28\times28}$\\
FC. 400  ReLU\\
FC. $L$, FC. $L$\\
\bottomrule
\end{tabular}
\end{center}
\end{minipage}
\begin{minipage}{.3\textwidth}
\begin{center}
\begin{tabular}{@{} l  @{}}
\toprule
\textbf{Decoder}\\ 
\midrule
Input $\in \mathbb{R}^{L}$\\
FC. 400  ReLU\\
FC. 1 x 28 x 28 Sigmoid\\
\bottomrule
\end{tabular}
\end{center}
\end{minipage}
\end{center}
\caption{Network architectures for MNIST-SVHN: MNIST.\label{tab:svhn_arch}}
\end{table}

\begin{table}[hbt!]
\begin{center}
\begin{tabular}{@{} l @{}}
\toprule
\textbf{Encoder}\\ 
\midrule
Input $\in \mathbb{R}^{3x32x32}$\\
 $4\times4$ conv. 32 stride 2 ReLU\\
 $4\times4$ conv. 64 stride 2 ReLU\\
 $4\times4$ conv. 128 stride 2 ReLU\\
 $4\times4$ conv. $L$ stride 1 ,  $4\times4$ conv. $L$ stride 1 \\
\bottomrule
\end{tabular}

\medskip

\begin{tabular}{@{} l@{}}
\toprule
\textbf{Decoder}\\ 
\midrule
Input $\in \mathbb{R}^{L}$ \\
 $4\times4$ upconv. 128 stride 1 ReLU\\
 $4\times4$ upconv. 64 stride 2 ReLU\\
 $4\times4$ upconv. 32 stride 2 ReLU\\
 $4\times4$ upconv. 3 stride 2 Sigmoid\\
\bottomrule
\end{tabular}

\end{center}
\caption{Network architectures for MNIST-SVHN: SVHN.\label{tab:mnist_arch}}
\end{table}

\paragraph{CUB-Captions.}  

The models were trained for 200 epochs with the learning rate $10^{-4}$. The best epoch was chosen based on the highest joint coherence evaluated on the validation set. We used a latent space dimensionality of $L=64$.
The network architectures are described in Table \ref{tab:cupcap_a} and Table~\ref{tab:cupcap_b} for the text and image modality, respectively.
\begin{table}[hbt!]
\begin{center}
\begin{tabular}{@{} l @{}}
\toprule
\textbf{Encoder}\\ 
\midrule
Input $\in \mathbb{R}^{1590}$\\
Word Emb. 256 \\
 $4\times4$ conv. 32 stride 2 BatchNorm2d  ReLU\\
 $4\times4$ conv. 64 stride 2 BatchNorm2d  ReLU\\
 $4\times4$ conv. 128 stride 2 BatchNorm2d  ReLU\\
 $1\times4$ conv. 256 stride  $1\times2$ pad  $0\times1$ \& BatchNorm2d  ReLU\\
 $1\times4$ conv. 512 stride  $1\times2$ pad  $0\times1$ \& BatchNorm2d  ReLU\\
 $4\times4$ conv. $L$ stride 1 ,  $4\times4$ conv. $L$ stride 1 \\
\bottomrule
\end{tabular}

\medskip

\begin{tabular}{@{} l @{}}
\toprule
\textbf{Decoder}\\ 
\midrule
Input $\in \mathbb{R}^{L}$\\
 $4\times4$ upconv. 512 stride 1 ReLU\\
 $1\times4$ upconv. 256 stride  $1\times2$ pad  $0\times1$ \& BatchNorm2d  ReLU\\
 $1\times4$ upconv. 128 stride  $1\times2$ pad  $0\times1$ \& BatchNorm2d  ReLU\\
 $4\times4$ upconv. 64 stride 2 BatchNorm2d  ReLU\\
 $4\times4$ upconv. 32 stride 2 BatchNorm2d  ReLU\\
 $4\times4$ upconv. 1 stride 2 ReLU\\
Word Emb.$^{\text{T}}$ 1590\\
\bottomrule
\end{tabular}

\end{center}
\caption{Network architectures for CUB-Captions language processing.\label{tab:cupcap_a}}
\end{table}

\begin{table}[hbt!]
\begin{center}
\begin{minipage}{.3\textwidth}
\begin{center}
\begin{tabular}{@{} l  @{}}
\toprule
\textbf{Encoder}\\ 
\midrule
Input $\in \mathbb{R}^{2048}$ \\
FC. 1024 ELU \\
FC. 512 ELU \\
FC. 256 ELU \\
FC. L, FC. L \\
\bottomrule
\end{tabular}
\end{center}
\end{minipage}
\begin{minipage}{.3\textwidth}
\begin{center}
\begin{tabular}{@{} l  @{}}
\toprule
\textbf{Decoder}\\ 
\midrule
 Input $\in \mathbb{R}^{L}$ \\
 FC. 256 ELU \\
 FC. 512 ELU \\
 FC. 1024 ELU \\
 FC. 2048 \\
\bottomrule
\end{tabular}
\end{center}
\end{minipage}
\end{center}
\caption{Network architectures for CUB-Captions image processing.\label{tab:cupcap_b}}
\end{table}

\paragraph{MNIST-Split-Three.}  

The models were trained for 50 epochs with the learning rate $2\cdot10^{-4}$. The best epoch was chosen by the highest accuracy of the reconstruction from the top half evaluated on the validation set.
We used a latent space dimensionality of $L=64$.  The network architectures are described in Table~\ref{tab:split_arch_three}.

\begin{table}[hbt!]
\begin{center}
\begin{minipage}{.49\textwidth}
\begin{center}
\begin{tabular}{@{} l @{}}
\toprule
\textbf{Encoder}\\ 
\midrule
Input $\in \mathbb{R}^{3x32x32}$ \\
 $4\times4$ conv. 64 stride 2 ReLU \\
 $4\times4$ conv. 128 stride 2 ReLU \\
 $4\times4$ conv. 256 stride 2 ReLU \\
FC. 786  ReLU \\
FC. L, FC. L \\
\bottomrule
\end{tabular}
\end{center}
\end{minipage}\begin{minipage}{.49\textwidth}
\begin{center}
\begin{tabular}{@{} l  @{}}
\toprule
\textbf{Decoder}\\ 
\midrule
Input $\in \mathbb{R}^{L}$ \\
FC. L  ReLU \\
FC. 512  ReLU \\
FC. 112  ReLU \\
 $4\times4$ upconv. 56 stride 1 ReLU \\
 $2\times4$ upconv. 28 stride 2 ReLU \\
\bottomrule
\end{tabular}
\end{center}
\end{minipage}
\end{center}
\caption{Network architectures for MNIST-Split-Three for each image part.\label{tab:split_arch_three}}
\end{table}

\paragraph{Spectra and molecule fingerprints.}  

The models were trained for 140 epochs with the learning rate $10^{-4}$. The best epoch was chosen by the highest accuracy of the test set.
We used a latent space dimensionality of $L=300$.  The network architectures are described in Table~\ref{tab:spectra} and Table~\ref{tab:fingerprints}.

\begin{table}[hbt!]
\begin{center}
\begin{minipage}{.49\textwidth}
\begin{center}
\begin{tabular}{@{} l @{}}
\toprule
\textbf{Encoder}\\ 
\midrule
Input $\in \mathbb{R}^{11000}$ \\
FC. 10000 \& BatchNorm1d  ReLU \\
FC. 5000 \& BatchNorm1d  ReLU  \\
FC. 2048 \& BatchNorm1d  ReLU  \\
FC. L, FC. L \\
\bottomrule
\end{tabular}
\end{center}
\end{minipage}\begin{minipage}{.49\textwidth}
\begin{center}
\begin{tabular}{@{} l  @{}}
\toprule
\textbf{Decoder}\\ 
\midrule
Input $\in \mathbb{R}^{L}$ \\
FC. 3000 \& BatchNorm1d  ReLU \\
FC. 5000 \& BatchNorm1d  ReLU  \\
FC. 10000 \& BatchNorm1d  ReLU  \\
FC. 11000 \\
\bottomrule
\end{tabular}
\end{center}
\end{minipage}
\end{center}
\caption{Network architectures for the spectra-fingerprints experiment, spectra modality.\label{tab:spectra}}
\end{table}

\begin{table}[hbt!]
\begin{center}
\begin{minipage}{.49\textwidth}
\begin{center}
\begin{tabular}{@{} l @{}}
\toprule
\textbf{Encoder}\\ 
\midrule
Input $\in \mathbb{R}^{2149}$ \\
FC. 1024  ReLU \\
FC. 1000  ReLU  \\
FC. L, FC. L \\
\bottomrule
\end{tabular}
\end{center}
\end{minipage}\begin{minipage}{.49\textwidth}
\begin{center}
\begin{tabular}{@{} l  @{}}
\toprule
\textbf{Decoder}\\ 
\midrule
Input $\in \mathbb{R}^{L}$ \\
FC. 500  ReLU \\
FC. 1024  ReLU  \\
FC. 2149 \& Sigmoid\\
\bottomrule
\end{tabular}
\end{center}
\end{minipage}
\end{center}
\caption{Network architectures for the spectra-fingerprints experiment, fingerprints modality.\label{tab:fingerprints}}
\end{table}






\end{document}

%% file: math_commands.tex
\usepackage{amsmath,amsfonts,bm}

\def\eqref#1{equation~\ref{#1}}

\def\1{\bm{1}}

\DeclareMathAlphabet{\mathsfit}{\encodingdefault}{\sfdefault}{m}{sl}
\SetMathAlphabet{\mathsfit}{bold}{\encodingdefault}{\sfdefault}{bx}{n}

\newcommand{\KL}{D_{\mathrm{KL}}}

